%% file: main.tex
\documentclass{article} 
\usepackage{iclr2027_conference,times}

\input{math_commands.tex}

\usepackage[utf8]{inputenc} 
\usepackage[T1]{fontenc}    
\usepackage{hyperref}       
\usepackage{url}            
\usepackage{booktabs}       
\usepackage{amsfonts}       
\usepackage{nicefrac}       
\usepackage{microtype}      
\usepackage{xcolor}         
\usepackage{graphicx}
\usepackage{amsmath}
\usepackage{amsthm}

\newtheorem{theorem}{Theorem}
\usepackage{algorithm}
\usepackage{algorithmic}
\usepackage{enumitem}       
\usepackage[many]{tcolorbox}

\newtcolorbox{abstractbox}{
  enhanced,
  boxrule=0pt,
  arc=0pt,
  outer arc=0pt,
  interior style={left color=blue!10, right color=red!10},
  fontupper=\small,
  before skip=10pt,
  after skip=10pt
}

\title{BoostLoRA: Growing Effective Rank by Boosting Adapters}

\author{%
  Raviteja Anantha \\
  Amazon \\
  \And
  Nick Levato \\
  Amazon \\
  \And
  Layne C.~Price \\
  Amazon \\
}

\iclrfinalcopy 
\begin{document}

\maketitle

\begin{abstract}
Parameter-efficient fine-tuning (PEFT) methods face a tradeoff between adapter
size and expressivity: ultra-low-parameter adapters are confined to
fixed low-rank subspaces, capping performance even with extended
training.
We propose BoostLoRA, a gradient-boosting framework that overcomes
this limit by iteratively training and merging minimal adapters on the
examples the current model gets wrong.
A \textsc{rotate} SVD basis strategy assigns each round to an
orthogonal subspace, so cumulative effective rank grows linearly with
the number of rounds while each adapter remains ultra-low-rank.
After merging, adapters are discarded, leaving zero inference overhead.
On Qwen2.5-3B, BoostLoRA reaches 89.1\% on GSM8K and 68.8\% on
MATH-500, surpassing both the best single-shot ultra-low-parameter adapter
(TinyLoRA) and full fine-tuning; on code generation it reaches 57.2\%
on MBPP and 80.4\% on HumanEval while full fine-tuning drops below the
zero-shot baseline.
We also demonstrate general applicability across domains via protein binding
classification with ESM2-650M and cross-entropy training.
BoostLoRA is, to our knowledge, the first ultra-low PEFT method whose effective
rank grows with training, separating per-round parameter cost from
total representational capacity.
\end{abstract}

\section{Introduction}
\label{sec:intro}
Parameter-efficient fine-tuning enables large language model adaptation without updating billions of weights.
LoRA~\citep{lora} injects trainable low-rank matrices into frozen layers; follow-up work has pushed the Pareto frontier further through shared random projections~\citep{vera}, SVD-informed initialization~\citep{pissa,adalora}, and weight tying~\citep{loraxs}.
TinyLoRA~\citep{tinylora} is the lowest-parameter point on this frontier, reducing trainable parameters to as few as one per module by projecting a low-dimensional vector through fixed random matrices within the SVD subspace of each weight matrix.
Combined with reinforcement learning (RL) \citep{grpo}, a 16-parameter TinyLoRA adapter improves Qwen2.5-3B from 76\% to 81\% on GSM8K, showing that RL is far more
parameter-efficient than supervised fine-tuning (SFT) at this scale.
However, individual ultra-low-parameter adapters hit an \emph{expressivity limit}~\citep{rlvr_peft}. Each adapter is confined to a fixed-rank subspace, and adding more parameters yields diminishing returns beyond a threshold. TinyLoRA's best configuration (8{,}064 parameters) plateaus at 87.2\% on GSM8K, within a point of full fine-tuning (87.0\%).

We propose \textbf{BoostLoRA}, a PEFT framework based on gradient boosting~\citep{friedman2001}. BoostLoRA trains $T$ sequential TinyLoRA adapters, each with ultra-low parameter budgets, on the examples the current model gets wrong. After each round, the adapter is merged into the base weights and discarded, leaving no inference overhead.
The core mechanism is a \textsc{rotate} SVD basis strategy that assigns each round's adapter to an orthogonal subspace. The cumulative weight update reaches effective rank $\min(rT, d, k)$ where each adapter is rank~$r$.
With $r = 2$ and $T = 20$ rounds, BoostLoRA builds the capacity of a rank-40 update while each round optimizes over a $2 \times 2$ \emph{transformation matrix}.

On GSM8K, BoostLoRA reaches \textbf{89.1\%}, beating TinyLoRA's best (87.2\% at 8{,}064 parameters) and full fine-tuning (87.0\% at 3.09B parameters).
On MATH-500 it reaches 68.8\%, and on code generation it achieves 57.2\% on MBPP and 80.4\% on HumanEval.
We also test BoostLoRA outside decoder models and generation: on PPB-Affinity~\citep{ppb_affinity} protein-protein binding classification with ESM2-650M~\citep{esm2} and cross-entropy training. The BoostLoRA framework improves over single-shot TinyLoRA throughout the ultra-low and mid-range parameter settings, while both degrade at the 129K-parameter setting.

We contribute:
\begin{enumerate}[leftmargin=*,itemsep=2pt]
  \item \textbf{BoostLoRA}, a gradient-boosted PEFT framework that
    resolves expressivity limits of ultra-low-parameter adapters
    by sequentially training and merging TinyLoRA adapters
    on failures (\S\ref{sec:method}).
  \item A \textbf{\textsc{rotate} SVD basis strategy} that ensures
    linear effective rank growth ($\leq rT$) across boosting rounds,
    with empirical confirmation that the bound is tight
    (\S\ref{sec:rank}, \S\ref{sec:basis_analysis}).
  \item An analysis of gradient isolation from correct examples
    (\S\ref{sec:boosting}), exact rank growth
    (Theorem~\ref{thm:rank}), and a non-vacuous generalization
    bound (Theorem~\ref{thm:gen}), which we validate empirically
    (App.~\ref{sec:theory_validation}).
  \item Results across math, code, and protein tasks on both decoder
    and encoder models, with both GRPO and cross-entropy training
    (\S\ref{sec:main_results}). 
\end{enumerate}

\section{Related Work}
\label{sec:related}
\paragraph{Parameter-efficient fine-tuning.} 
LoRA~\citep{lora} adapts frozen weights with low-rank updates. AdaLoRA~\citep{adalora}
allocates rank by importance, LoRA-XS~\citep{loraxs} learns a small core between
frozen SVD factors, VeRA~\citep{vera} shares random projections, DoRA~\citep{dora}
separates magnitude and direction, PiSSA~\citep{pissa} initializes principal SVD
directions, and TinyLoRA~\citep{tinylora} uses a trainable vector in a fixed SVD
subspace. These methods use a single adapter whose update rank is fixed at
creation; BoostLoRA instead accumulates disjoint low-rank updates across rounds.

\paragraph{Iterative and boosted adaptation.}
CoLA~\citep{cola} is the closest prior method that sequentially merges LoRA adapters, and we compare against it directly in Appendix~\ref{sec:cola_comparison}. In our matched implementation, CoLA accuracy generally peaks within the first three merges and does not improve consistently with chain length (Table~\ref{tab:chain-of-lora}). Its configurations remain below BoostLoRA, suggesting that iterative merging alone saturates without explicitly disjoint subspaces.

MELoRA~\citep{melora} increases effective rank by stacking mini-LoRA modules within one training round. Model Soups~\citep{modelsoups} and LoRA Soups~\citep{lorasoups} average independently trained adapters for skill composition rather than sequential residual fitting. ReLoRA~\citep{relora} periodically merges and restarts standard-rank LoRA during pre-training, but uses millions of parameters per round, trains on the full dataset, and does not enforce orthogonality across rounds.

XGBLoRA~\citep{xgblora} also applies gradient boosting to rank-1 LoRA adapters. However, its adapters use approximately 2.4M parameters per round on LLaMA3-8B, compared with BoostLoRA's 12 parameters; it uses SFT, which TinyLoRA~\citep{tinylora} found substantially less parameter-efficient than RL at scale; and it evaluates commonsense reasoning rather than math or code.

HRA~\citep{NEURIPS2024_cdd06402} parameterizes each frozen weight matrix with a chain of learnable Householder reflections, but is a single-run method rather than sequential failure-set adaptation. O-LoRA~\citep{wang-etal-2023-orthogonal} keeps low-rank subspaces orthogonal across tasks to reduce forgetting. In contrast, BoostLoRA assigns disjoint SVD blocks across rounds of the same task, using orthogonality to grow within-task capacity.

\paragraph{RL for reasoning.}
GRPO~\citep{grpo} estimates advantages from within-group reward statistics without a PPO-style critic. DeepSeek R1~\citep{deepseekr1}, SimpleRL~\citep{simplerl}, and RLOO~\citep{rloo} demonstrate that pure or simplified RL can produce strong reasoning behavior without an SFT stage or complex PPO variants. Under these dynamics, ultra-low-parameter adapters face an \emph{expressivity limit}: VeRA and rank-1 LoRA fail to learn reasoning, while SVD-based initializations such as PiSSA can undergo spectral collapse~\citep{rlvr_peft}. BoostLoRA addresses this limit by accumulating effective rank across sequential ultra-low-parameter adapters.

\paragraph{Gradient boosting theory.}
Gradient boosting~\citep{friedman2001} greedily adds weak learners that fit the current residual. Its margin theory~\citep{adaboost,schapire1997margin} relates generalization to the margin distribution rather than ensemble size. XGBoost \citep{xgboost} scaled this principle to large-scale tree learning, while BoostResNet~\citep{boostresnet} established exponentially decreasing training error for sequential residual blocks under a weak-learning condition. BoostLoRA applies this framework to PEFT: TinyLoRA adapters are weak learners, and \textsc{rotate} grows their cumulative capacity linearly across rounds.

\paragraph{Protein language models and binding affinity.}
ESM2~\citep{esm2} is a protein language model family trained on evolutionary sequences. PPB-Affinity~\citep{ppb_affinity} aggregates protein-protein binding affinities from structural databases and has primarily been studied as a regression task (best Pearson $r=0.701$~\citep{ppb_affinity}; Spearman $\rho=0.48$~\citep{plm_ppi}). To evaluate BoostLoRA on an encoder-only model with cross-entropy training, we binarize the task at $K_D<500$\,nM.

\section{Method}
\label{sec:method}
BoostLoRA repeatedly trains and merges minimal adapters with ultra-low parameter budgets to incrementally improve a frozen underlying model.
Each round (i)~identifies the examples the current model gets wrong, (ii)~trains a fresh TinyLoRA adapter on that failure set via GRPO (for generation tasks) or cross-entropy (for classification), and (iii)~merges the adapter into the base weights.

\subsection{TinyLoRA adapter parameterization}
\label{sec:adapter}
For each linear layer $W$ that is $d \times k$ dimensions
in the base model, we compute a rank-$r$ truncated SVD, by $W \approx U \Sigma V^\top$. 
The adapter introduces a weight update 
\begin{equation}
  \Delta W = U \, \Sigma \, R \, V^\top,
  \qquad
  R = \sum_{i=1}^{u} v_i \, P_i,
  \label{eq:adapter}
\end{equation}
where $P_i$ are fixed $r \times r$ random projection matrices, and $v \in \mathbb{R}^{u}$ is the \emph{only} trainable vector. $v$ selects a linear combination of random rotations within the rank-$r$ subspace of $W$, modulated by the singular values~$\Sigma$.

Following~\citet{tinylora}, we use block-level weight tying: the adapted modules (e.g., 252 for Qwen2.5-3B, 198 for ESM2-650M) are partitioned into $g$ tying groups using a tiled assignment, and all modules within a group share the same~$v$. With $g = 4$ groups and projection dimension $u = 3$, the total number of trainable parameters per boosting round is $g \times u = 12$.

\subsection{Gradient boosting with TinyLoRA weak learners}
\label{sec:boosting}
Algorithm~\ref{alg:boostlora} summarizes the procedure. Let $W_0$ denote the original pretrained weights. At round $t$:

\begin{enumerate}[leftmargin=*,itemsep=2pt]
  \item \textbf{Evaluate.} Run the current model $\mathcal{M}$ on the full training
        set and collect the \emph{failure set},
        $\mathcal{F}_t = \{(x, y) : \mathcal{M}(x) \neq y\}$.
  \item \textbf{Train.} Initialize a fresh TinyLoRA adapter by setting $v \leftarrow 0$
        and train it via GRPO or cross-entropy on
        $\mathcal{F}_t$.
  \item \textbf{Merge.} Compute $\Delta_t = U \Sigma R_t V^\top$ and
        fold it into the base weights,
        $W_{t} \leftarrow W_{t-1} + \Delta_t$.
  \item \textbf{Accumulate.} Track the cumulative update for effective rank measurement:
        $\bar{\Delta}_t \leftarrow \bar{\Delta}_{t-1} + \Delta_t$.
\end{enumerate}

\noindent
After merging, the adapter is discarded and the model runs at its original inference cost, with no adapter overhead. This parallels gradient boosting~\citep{friedman2001}, where each TinyLoRA adapter is a weak learner that fits the ensemble's current residual (the failure set).

\paragraph{Dynamic residuals.} Merging $\Delta_t$ changes the underlying model and can both correct previous failures and introduce new ones. Consequently, we recompute $\mathcal{F}_t$ from the current model at every round. BoostLoRA follows the stage-wise additive structure of gradient boosting, while its residual is defined by the round's task-level correctness rather than a fixed output-space residual.



\paragraph{Gradient isolation.}
Each round trains only on $\mathcal{F}_t$, so its gradient contains no direct contribution from currently correct examples. Merging can nevertheless change any prediction: a correct example regresses only if the update overcomes its logit margin. Empirically, the small per-round updates ($\|\Delta W_t\|_F \approx 0.01$) cause approximately 2\% regression of previously correct examples per round.

\begin{algorithm}[t]
\caption{BoostLoRA: Gradient-boosted TinyLoRA}
\label{alg:boostlora}
\begin{algorithmic}[1]
\REQUIRE Pretrained model $\mathcal{M}$ with weights $W_0$, training set
$\mathcal{D}$, SVD rank $r$, rounds $T$, basis strategy $\in \{\textsc{top}, \textsc{rotate}\}$
\STATE Initialize cumulative delta $\bar{\Delta} \leftarrow 0$
\FOR{$t = 1$ \TO $T$}
  \STATE $\mathcal{F}_t \leftarrow \{(x,y) \in \mathcal{D} : \mathcal{M}(x) \neq y\}$
        \hfill $\triangleright$ Identify failures
  \IF{$|\mathcal{F}_t| < $ threshold}
    \STATE \textbf{break} \hfill $\triangleright$ Early stop
  \ENDIF
  \STATE Select SVD components $(U_t, \Sigma_t, V_t)$ using basis strategy and round $t$
  \STATE $v_t \leftarrow \text{GRPO}(\mathcal{M}, \mathcal{F}_t, U_t, \Sigma_t, V_t)$
        \hfill $\triangleright$ Train adapter
  \STATE $\Delta_t \leftarrow U_t \Sigma_t R(v_t) V_t^\top$
        \hfill $\triangleright$ Compute weight update
  \STATE $W_t \leftarrow W_{t-1} + \Delta_t$
        \hfill $\triangleright$ Merge into base
  \STATE $\bar{\Delta} \leftarrow \bar{\Delta} + \Delta_t$
        \hfill $\triangleright$ Track cumulative change
\ENDFOR
\RETURN $\mathcal{M}$ with weights $W_T$
\end{algorithmic}
\end{algorithm}

\subsection{SVD basis strategies and effective rank growth}
\label{sec:rank}
The choice of which singular vectors to use in each round determines the \emph{effective rank} of the cumulative weight update $\sum_t \Delta_t$. We consider two strategies:
\begin{enumerate}
\item  \textbf{Top basis.}
Every round computes the top-$r$ singular vectors of the current weight matrix $W_{t-1}$. Because the adapter norm $|\Delta_t|_F \approx 0.005$--$0.013$ is small relative to $|W|_F$, the top singular vectors change minimally between rounds. Successive adapters therefore operate in the same rank-$r$ subspace, and the cumulative update remains approximately rank~$r$ regardless of~$T$.

\item \textbf{Rotate basis.}
Before training, we compute the full SVD on the pretrained weight matrix $W_0$ once. Round $t$ then uses singular vectors $r(t{-}1){+}1$ through $rt$, corresponding to the SVD offset $r \cdot (t{-}1)$. Because all rounds index disjoint blocks from the same fixed SVD, their left and right subspaces are mutually orthogonal. This ensures that
\begin{equation}
  \mathrm{rank}\!\Bigl(\textstyle\sum_{t=1}^{T} \Delta_t\Bigr)
  \;=\; \min(rT,\; d, \; k).
  \label{eq:rankgrowth}
\end{equation}
Caching the SVD outside the boosting loop produced final performance within the same confidence interval and preserved an $\varepsilon$-rank of 40. Rollout generation dominates runtime, this optimization reduced wall-clock time by only 0.3-0.6\%.
\end{enumerate}

\paragraph{Effective rank measurement.}
We track the cumulative delta $\bar{\Delta}_t = \sum_{s=1}^{t} \Delta_s$ in float32 (to avoid bf16 rounding noise that inflates apparent rank) throughout training and report two rank measures:
(i)~the \emph{participation ratio} $\rho = (\sum_i \sigma_i)^2 / \sum_i \sigma_i^2$,
which equals 1 for a rank-1 matrix and $\min(d,k)$ for uniform singular values $\sigma_i$; and 
(ii)~the \emph{$\varepsilon$-rank}, the count of singular values exceeding $\varepsilon \cdot \sigma_1$ with $\varepsilon = 0.01$.

\subsection{Training procedure}
\label{sec:grpo}
Each round trains the adapter using GRPO~\citep{grpo} for generation tasks or cross-entropy for classification tasks. For GRPO, we sample a group of $G = 8$ completions per prompt $x$ in the failure set, score them with a binary reward function (exact-match for math, sandboxed execution for code), and compute advantages via within-group normalization:
\begin{equation}
  \hat{A}_i = \frac{r_i - \bar{r}}{\sigma_r + \epsilon},
  \qquad
  \bar{r} = \frac{1}{G}\sum_{j=1}^{G} r_j.
  \label{eq:advantage}
\end{equation}
The policy gradient uses clipped importance weights with $\epsilon_c = 0.2$, plus an optional KL penalty toward the reference policy. We use AdamW with cosine learning rate (LR) scheduling (warmup ratio 0.1) and gradient clipping at 1.0, training for three epochs per round.

\subsection{Theoretical properties}
\label{sec:theory}
We establish two formal results: the \textsc{rotate} basis achieves exact rank growth (Theorem~\ref{thm:rank}), and the generalization penalty from $T$ rounds is bounded by the total adapter norm, which grows sublinearly due to self-limiting dynamics (Theorem~\ref{thm:gen}).

\begin{theorem}[Exact Rank Growth]\label{thm:rank}
For BoostLoRA with \textsc{rotate} basis, SVD rank $r$, and $T$ rounds ($rT \leq \min(d,k)$), if each round's mixing matrix $R_t = \sum_i (v_t)_i P_i$ has rank~$r$, the cumulative rank is:
$\mathrm{rank}\bigl(\sum_{t=1}^{T} \Delta_t\bigr) = rT$.
\end{theorem}
\begin{proof}
Round $t$ uses left singular vectors $r(t{-}1){+}1$ through $rt$ as columns of $U_t$, and the corresponding right singular vectors as columns of $V_t$, from the full SVD of $W_0$. Since $\Delta_t = U_t \Sigma_t R_t V_t^\top$, we have $\mathrm{col}(\Delta_t) \subseteq \mathrm{col}(U_t)$ and $\mathrm{row}(\Delta_t) \subseteq \mathrm{col}(V_t)$. Because $U_t$ and $V_t$ use distinct columns of orthogonal matrices, both column and row spaces are mutually orthogonal across rounds, so ranks add: $\mathrm{rank}(\sum_t \Delta_t) = \sum_t \mathrm{rank}(\Delta_t) = rT$.
\end{proof}

With \textsc{top} basis, all rounds reuse the same columns $U_1$, so $\mathrm{rank}(\sum_t \Delta_t) \leq r$ regardless of $T$. The empirical $\varepsilon$-rank in Figure~\ref{fig:rank_comparison} confirms this, showing \textsc{rotate} reaches rank 40 at round~20 while \textsc{top} stays at $\approx$2.

To prove generalization bounds, for each input $x$, choose a finite, fixed evaluation set of possible outputs $\mathcal{Y}(x) = \mathcal{Y}^{+}(x) \cup \mathcal{Y}^{-}(x)$, with $\mathcal{Y}^{+}(x)$ the outputs the task's reward marks correct and $\mathcal{Y}^{-}(x)$ those it marks incorrect. Let $s_T(x,y)$ be the model's log-probability of output $y$ after $T$ rounds. The margin 
$m_T(x) = \max_{y \in \mathcal{Y}^{+}} s_T(x,y) - \max_{y \in \mathcal{Y}^{-}} s_T(x,y)$
is the gap between the best correct output and the best incorrect one. Define a model error when $m_T(x) \leq 0$.
Merging round $t$'s adapter perturbs each adapted layer's output by $\Delta W_\ell\, h_\ell(x)$, where $h_\ell(x)$ is the input to layer~$\ell$. To first order, the resulting change in margin is $\Delta m(x) \approx \sum_\ell g_\ell(x)^\top \Delta W_\ell\, h_\ell(x)$, where $g_\ell(x)$ is the gradient of the margin with respect to layer~$\ell$'s output. Substituting $\Delta W_\ell = U_\ell \Sigma_\ell R_t V_\ell^\top$ with $R_t = \sum_i (v_t)_i P_i$ from Eq.~\ref{eq:adapter} gives $\Delta m(x) \approx \sum_i (v_t)_i \sum_\ell g_\ell(x)^\top U_\ell \Sigma_\ell P_i V_\ell^\top h_\ell(x) = \langle v_t, \phi_t(x) \rangle$, which defines the feature map $\phi_t(x) \in \mathbb{R}^u$. Each round is thus a linear predictor in its trainable vector $v_t$. Write $\|v_t\|_2 \leq B_t$, $B_{\mathrm{total}} = \sum_t B_t$, and $X = \max_{t,x} \|\phi_t(x)\|_2$.

\begin{theorem}[Generalization Bound]\label{thm:gen}
For any $\theta > 0$ and $\delta > 0$, with probability $\geq 1 - \delta$ over a training set of size $n$,
\begin{equation}
\Pr_{\mathrm{test}}[\mathrm{error}] \;\leq\; \Pr_{\mathrm{train}}[m_T \leq \theta] \;+\;  \frac{ 2 X B_{\mathrm{total}}  }{\theta \sqrt{n}} \;+\; \sqrt{\frac{\log(2/\delta)}{2n}} .
\label{eq:gen}
\end{equation}
\end{theorem}

The proof in Appendix~\ref{sec:gen_proof} relies on each round's adapter being a linear function of $v_t$, as shown above, and bounds the Rademacher complexity of the ensemble via subadditivity.
The bound depends on $B_{\mathrm{total}}$, not $T$ directly. Because adapter norms decay as failures shrink (Figure~\ref{fig:dynamics}), $B_{\mathrm{total}}$ grows sublinearly in $T$, so additional rounds do not significantly hurt generalization.
Under \textsc{rotate} the orthogonal per-round blocks tighten the complexity term by up to $\sqrt{T}$ relative to \textsc{top}. See Appendix~\ref{sec:gen_proof} for a proof and Appendix~\ref{sec:theory_validation} for numerical evaluation.

\section{Experiments}
\label{sec:experiments}

\paragraph{Models.}
Math and code experiments use Qwen2.5-3B-Instruct~\citep{qwen2.5}, a 36-layer decoder with hidden dimension 2{,}048 and grouped-query attention (16 query heads, 2 KV heads). All 252 linear projections are adapted. Protein experiments use ESM2-650M~\citep{esm2}, a 33-layer encoder with 1{,}280 hidden dimension trained on evolutionary protein sequences; all 198 linear projections are adapted.
Unless otherwise noted: SVD rank $r = 2$, $g = 4$ tying groups, projection dimension $u = 3$, yielding 12 trainable parameters per adapter per round for $T = 20$ boosting rounds.

\paragraph{Benchmarks and baselines.}
We evaluate on five benchmarks across three domains:
(1) \emph{Math}: GSM8K~\citep{gsm8k} and MATH-500~\citep{math500}; 
(2) \emph{Code}: MBPP~\citep{mbpp} for training and evaluation, and HumanEval~\citep{humaneval} as a held-out cross-benchmark, following the protocol of~\citet{stepcoder,mucode};
and
(3) \emph{Proteins}: PPB-Affinity~\citep{ppb_affinity}, a protein-protein binding dataset. We formulate it as binary classification using the common threshold $K_D < 500$\,nM to denote binding.

We compare against four baselines:
(i)~the pretrained base model (zero-shot),
(ii)~TinyLoRA~\citep{tinylora} at 16, 252, 8{,}064, and 129{,}024 parameters (single adapter, single round),
(iii)~full fine-tuning at 3.09B parameters for Qwen-Instruct and 651M for ESM2,
and (iv)~for protein only, a frozen-ESM2 linear probe (logistic regression on mean-pooled embeddings, 1{,}281 parameters).
Math baselines are taken from~\citet{tinylora}.

\paragraph{Ablations.}
We study three axes:
(i)~SVD basis (\textsc{top} vs.\ \textsc{rotate}) on GSM8K and MATH-500, shown in Table~\ref{tab:ablation};
(ii)~a monolithic rank-40 adapter ($r{=}40$, $T{=}1$) that matches \textsc{rotate}'s total subspace in one round;
(iii)~parameter scaling on PPB-Affinity for both TinyLoRA and BoostLoRA at 12 to 129{,}024 parameters, shown in Table~\ref{tab:protein}.

\paragraph{Training details.}
For math and code, we use GRPO with group size $G = 8$, learning rate $5 \times 10^{-4}$, cosine schedule, and 3 epochs per round.
For protein modeling, we use cross-entropy with two-phase training per round. Phase~1 trains the classification head on the full dataset. Phase~2 trains the adapter on the failure set with the head frozen.
Models run in bf16 throughout.
All experiments use 8 GPUs with distributed data parallelism (DDP).
Accuracy figures include 95\% bootstrap confidence intervals.

\paragraph{Learning rate and stopping.}
We use a fixed learning rate of $5\times10^{-4}$ for the default 12-parameter adapter. For a per-round parameter budget $p>12$, we use
\begin{equation}
    \eta(p) = 5\times10^{-4}\sqrt{\frac{12}{p}}.
\end{equation}
This scaling limits the magnitude of increasingly expressive adapters. We additionally monitor $\|v_t\|_2$: failure of the adapter norm to decrease as the failure set shrinks indicates that later learners may be too strong.

Training terminates when either
(i)~the failure set contains fewer than eight examples for math and protein tasks, or fewer than sixteen examples for MBPP, or
(ii)~accuracy fails to improve for five consecutive rounds.
The second criterion is applied only after completing at least five rounds. These rules were fixed across the reported experiments rather than tuned independently for each dataset. Detailed training costs are reported in Appendix~\ref{sec:training_cost}.

\section{Main results}
\label{sec:main_results}
\begin{table}[t]
    \centering
    \caption{\textbf{Benchmark results on Qwen2.5-3B-Instruct.}
    TinyLoRA baselines trained with GRPO~\citep{tinylora};
    math baselines from~\citet{tinylora};
    code baselines are our own TinyLoRA runs using the same codebase.
    $^\dagger$Full fine-tuning (math from~\citet{tinylora}).
    $^*$Full FT degrades HumanEval from 72.6\% to 57.9\%, matching
    narrow-dataset SFT degradation reported by~\citet{stepcoder}.
    Parameter counts denote the maximum number of adapter parameters optimized simultaneously. BoostLoRA trains 12 parameters per round, optimizes $20\times12=240$ adapter coordinates over the complete sequence, and retains zero additional parameters after merging.}
    \label{tab:results}
    \small
    \setlength{\tabcolsep}{5pt}
    \begin{tabular}{lc cc cc}
        \toprule
        & & \multicolumn{2}{c}{\textbf{Mathematical Reasoning}}
        & \multicolumn{2}{c}{\textbf{Code Generation}} \\
        \cmidrule(lr){3-4} \cmidrule(lr){5-6}
        \textbf{\shortstack{~\\Method}}
        & \textbf{\shortstack{Peak Trainable\\Adapter Params}} 
        & GSM8K & MATH-500 & MBPP & HumanEval \\
        \midrule
        Base model (zero-shot)  & 0              & 76.0  & 55.0   & 49.8   & 72.6   \\
        \midrule
        TinyLoRA                & 16             & 80.9  & 64.0   & 50.6   & 63.4  \\
        TinyLoRA                & 252            & 85.4  & 66.4   & 52.2   & 64.0  \\
        TinyLoRA                & 8{,}064        & 87.2  & 67.8   & 52.6 & 64.6  \\
        TinyLoRA                & 129{,}024      & 86.7  & 67.8   & 54.4 & 67.7  \\
        \midrule
        Full FT$^\dagger$       & 3.09B          & 87.0  & 69.0   & 50.4 & 57.9$^*$  \\
        \midrule
        \textbf{BoostLoRA (ours)} & \textbf{12}
          & \textbf{89.1} & \textbf{68.8} & \textbf{57.2} & \textbf{80.4} \\
        \bottomrule
    \end{tabular}
\end{table}

\begin{figure}[t]
    \label{fig:math500}
    \centering
    \includegraphics[width=0.65\linewidth]{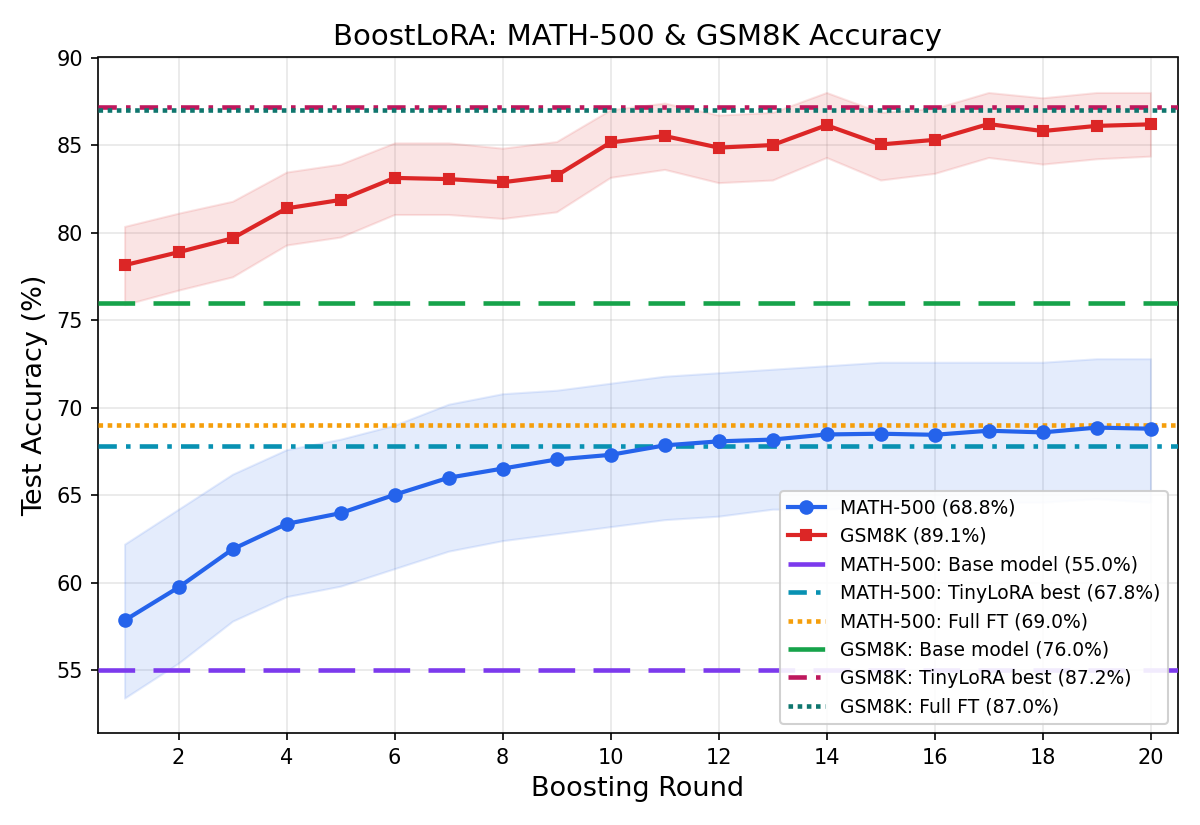}
    \caption{\textbf{Dual evaluation:} MATH-500
    (blue) and GSM8K (red) improve concurrently
    when training on SimpleRL~\citep{simplerl},
    suggesting general reasoning gains.
    MATH-500 beats TinyLoRA best at round~12 and saturates near Full FT at round~14.}
    \label{fig:math_dual_eval}
\end{figure}

\paragraph{Mathematical reasoning.}
Table~\ref{tab:results} shows that BoostLoRA reaches \textbf{89.1\%} on GSM8K, beating TinyLoRA's best at 8{,}064 parameters (87.2\%) by 1.9 points and full fine-tuning at 3.09B parameters (87.0\%) by 2.1 points.
Figure~\ref{fig:math_dual_eval} shows BoostLoRA reaching \textbf{68.8\%} on MATH-500, exceeding TinyLoRA's best (67.8\%) and approaching full fine-tuning (69.0\%).
The slower saturation relative to GSM8K (round~14 vs.\ round~12) reflects the greater difficulty of competition-level problems. The dual evaluation (Figure~\ref{fig:math_dual_eval}) shows concurrent improvement on both MATH-500 and GSM8K (76.0\% $\to$ 89.1\%) when training on SimpleRL~\citep{simplerl}.

\paragraph{Code generation.}
Figure~\ref{fig:gsm8k_deep_dive}(c) presents results when training on the MBPP train split (374~problems) with code-execution reward. MBPP accuracy improves from 49.8\% to \textbf{57.2\%} over 20 rounds
(15\% relative improvement). HumanEval, evaluated as a held-out cross-benchmark, peaks at \textbf{80.4\%} and stabilizes around 79\%. The cross-benchmark transfer suggests that BoostLoRA learns general coding ability rather than MBPP-specific patterns.
By contrast, full fine-tuning \emph{degrades} HumanEval from 72.6\% to 57.9\%. This matches prior findings (\citet{stepcoder}) where fine-tuning DeepSeek-Coder on APPS+ caused a similar performance drop (HumanEval: 78.0\% $\to$ 55.5\%), attributing degraded generalization from SFT on the narrow-dataset.

\begin{table*}[t]
\centering
\caption{\textbf{PPB-Affinity binding classification (ESM2-650M).}
Binary classification at $K_D = 500$\,nM threshold.
TinyLoRA: single round, trained on full dataset.
BoostLoRA: 20 rounds with \textsc{rotate} basis, failure-focused
training, LR scaled as $5{\times}10^{-4} \cdot \sqrt{12/p}$ for
$p > 12$.
No prior classification baselines exist for this dataset.
$^\ddagger$TinyLoRA and BoostLoRA both struggle at 129K params: large adapters disrupt ESM2's pretrained representations (see text).}
\label{tab:protein}
\small
\setlength{\tabcolsep}{4pt}
\begin{tabular}{l cc ccccc ccccc}
\toprule
\textbf{Metric}
  & \textbf{Linear} & \textbf{Full}
  & \multicolumn{5}{c}{\textbf{TinyLoRA}}
  & \multicolumn{5}{c}{\textbf{BoostLoRA (ours)}} \\
  & \textbf{Probe} & \textbf{FT} & & & & & & & & & & \\
\cmidrule(lr){4-8} \cmidrule(lr){9-13}
\textit{Params}
  & \textit{1{,}281} & \textit{651M}
  & 12 & 240 & 4{,}032 & 8{,}064 & 129K
  & 12 & 240 & 4{,}032 & 8{,}064 & 129K \\
\midrule
Acc (\%) & 59.7 & 69.4 & 66.3 & 64.3 & 65.1 & 65.3 & 65.3 & \textbf{67.9} & 68.5 & \textbf{69.1} & 68.1 & 58.2 \\
F1       & 76.0 & 81.0 & 80.4 & 80.5 & 79.3 & 79.0 & 79.0 & 80.1 & 80.6 & \textbf{81.0} & 80.1 & 79.2 \\
AUC      & 58.5 & 67.0 & 68.0 & 68.0 & 60.5 & 56.2 & 50.6$^\ddagger$ & 67.7 & 68.4 & \textbf{69.0} & 65.8 & 48.5$^\ddagger$ \\
\bottomrule
\end{tabular}
\end{table*}

\paragraph{Protein modeling.}
Table~\ref{tab:protein} reports accuracy, F1, and AUC-ROC across parameter scales for both TinyLoRA (single round) and BoostLoRA (20 rounds) for PPB-Affinity. Boosting helps most at ultra-low parameters: at 12 parameters, BoostLoRA (67.9\%) beats TinyLoRA (66.3\%) by 1.6 points. We observe that TinyLoRA degrades with more parameters, increasing from 12 to 129{,}024 parameters \emph{reduces} AUC from 68\% to 50.6\% (near random).
The linear probe (frozen ESM2 + logistic regression, 1{,}281 parameters) reaches only 59.7\% accuracy with AUC = 0.585. BoostLoRA at 12 parameters per round closes most of the gap to full fine-tuning (651M parameters, 69.4\%).
At 129{,}024 parameters, both methods struggle, with TinyLoRA's AUC dropping to 0.506 and BoostLoRA to 0.485 (random).

\begin{figure}[t]
    \centering
    \includegraphics[width=\linewidth]{results/comparison_figures/gsm8k_deep_dive_new}
    \caption{\textbf{BoostLoRA across domains.}
    \textbf{(a)}~GSM8K: \textsc{top} (red) and \textsc{rotate} (blue)
    track identically through round~8, then diverge: \textsc{top}
    saturates at 87.7\% while \textsc{rotate} reaches 89.1\%.
    \textbf{(b)}~GSM8K method comparison: BoostLoRA \textsc{rotate}
    (89.1\%) beats TinyLoRA at all param counts, full FT (87.0\%),
    and the $r{=}40$ monolithic ablation (85.2\%).
    \textbf{(c)}~Code generation: MBPP (blue) improves from 49.8\% to
    57.2\%; HumanEval (red, cross-benchmark) peaks at 80.4\%.}
    \label{fig:gsm8k_deep_dive}
\end{figure}

\begin{table}[t]
\centering
\caption{\textbf{Ablation: SVD basis strategy and rank.}
All BoostLoRA variants use 12 trainable parameters per adapter. The $r{=}40$ ablation matches \textsc{rotate}'s total subspace in a single round.}
\label{tab:ablation}
\small
\begin{tabular}{lcccc}
\toprule
\textbf{Method} & \textbf{Params/Adapter}
  & \textbf{Rounds ($T$)} & \textbf{GSM8K} & \textbf{MATH-500} \\
\midrule
Base model (zero-shot)              & ---  & ---  & 76.0 & 55.0 \\
TinyLoRA                            & 12   & 1    & 80.9 & 64.0 \\
TinyLoRA                            & 240  & 1    & 85.4 & 66.4 \\
\midrule
BoostLoRA ($r{=}40$, ablation)     & 12   & 1    & 85.2 & 64.8 \\
BoostLoRA \textsc{top} ($r{=}2$)   & 12   & 20   & 87.7 & 67.3 \\
\textbf{BoostLoRA \textsc{rotate} ($r{=}2$)} & \textbf{12} & \textbf{20}
  & \textbf{89.1} & \textbf{68.8} \\
\bottomrule
\end{tabular}
\end{table}

\begin{figure}[t]
    \centering
    \includegraphics[width=\linewidth]{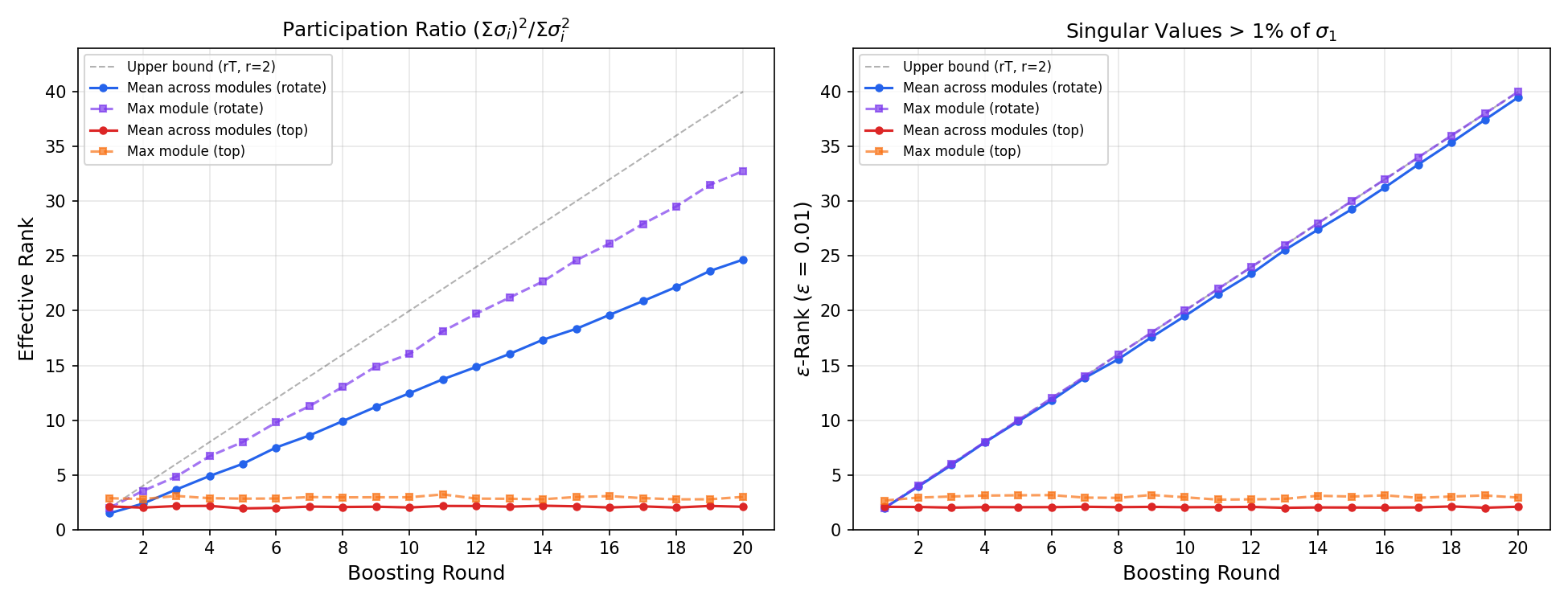}
    \caption{\textbf{Effective rank of cumulative weight update
    $\bar{\Delta} = \sum_t \Delta_t$.}
    Left: participation ratio. Right: $\varepsilon$-rank
    ($\varepsilon = 0.01$).
    \textsc{Top} (red): rank stays flat at ${\approx}\,2$.
    \textsc{Rotate} (blue): rank grows linearly to ${\approx}\,25$
    (participation ratio) and ${\approx}\,40$
    ($\varepsilon$-rank), matching $rT = 40$.}
    \label{fig:rank_comparison}
\end{figure}

\paragraph{Effective rank growth with \textsc{rotate}.}
\label{sec:basis_analysis}
Figure~\ref{fig:gsm8k_deep_dive}(a) shows that \textsc{top} and \textsc{rotate} produce identical accuracy trajectories through round~8.
After round~8, \textsc{top} saturates at 87.7\% while \textsc{rotate} continues to 89.1\%, a 1.4-point gap. This divergence aligns with the rank-capacity limit: by round~8, \textsc{top} has exhausted its fixed rank-2 subspace, while \textsc{rotate} has accumulated rank~16 and continues adding orthogonal directions. The tight match between observed $\varepsilon$-rank and the bound $rT$ confirms that \textsc{rotate} produces rank growth that translates to accuracy gains.
Figure~\ref{fig:rank_comparison} validates~\eqref{eq:rankgrowth}.
With \textsc{rotate}, $\varepsilon$-rank grows linearly to 40 = $rT$, while the participation ratio reaches $\approx$25 (below the bound because early-round adapters contribute larger $\|\Delta_t\|_F$, creating non-uniform singular values).
This rank gap explains the 1.4-point accuracy difference on GSM8K (Table~\ref{tab:ablation}) and motivates \textsc{rotate} as the default.

\paragraph{Rank-2 and monolithic rank-40.} Figure~\ref{fig:gsm8k_deep_dive}(b) compares all methods. The monolithic $r = 40$, $T = 1$ adapter, which matches \textsc{rotate}'s total subspace in one round, reaches only 85.2\%, \textbf{3.9 points below} boosted \textsc{rotate}. With $r = 2$, each of the 12 parameters controls $\nicefrac{1}{4}$ of the entries in a $2 \times 2$ matrix $R$, whereas with $r = 40$ the same 12 parameters must influence a $40 \times 40$ matrix, diluting the gradient signal.

\begin{figure}[!b]
    \centering
    \includegraphics[width=0.48\linewidth]{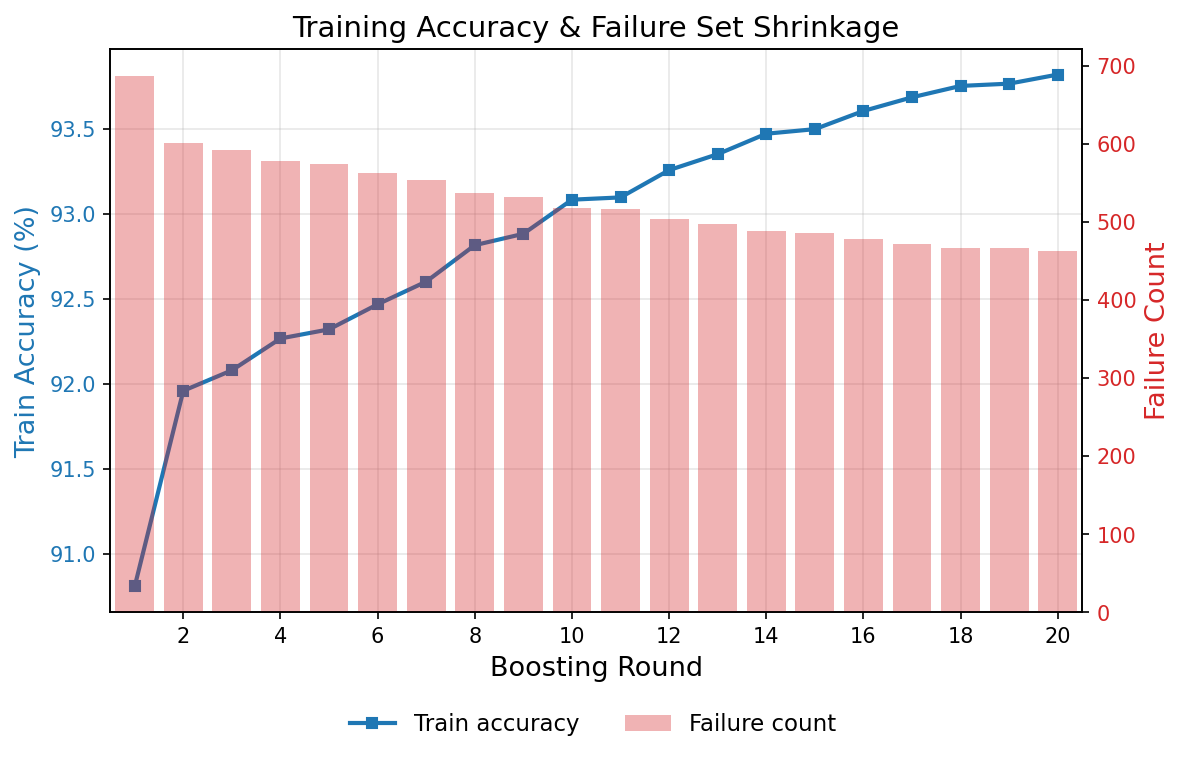}
    \hfill
    \includegraphics[width=0.48\linewidth]{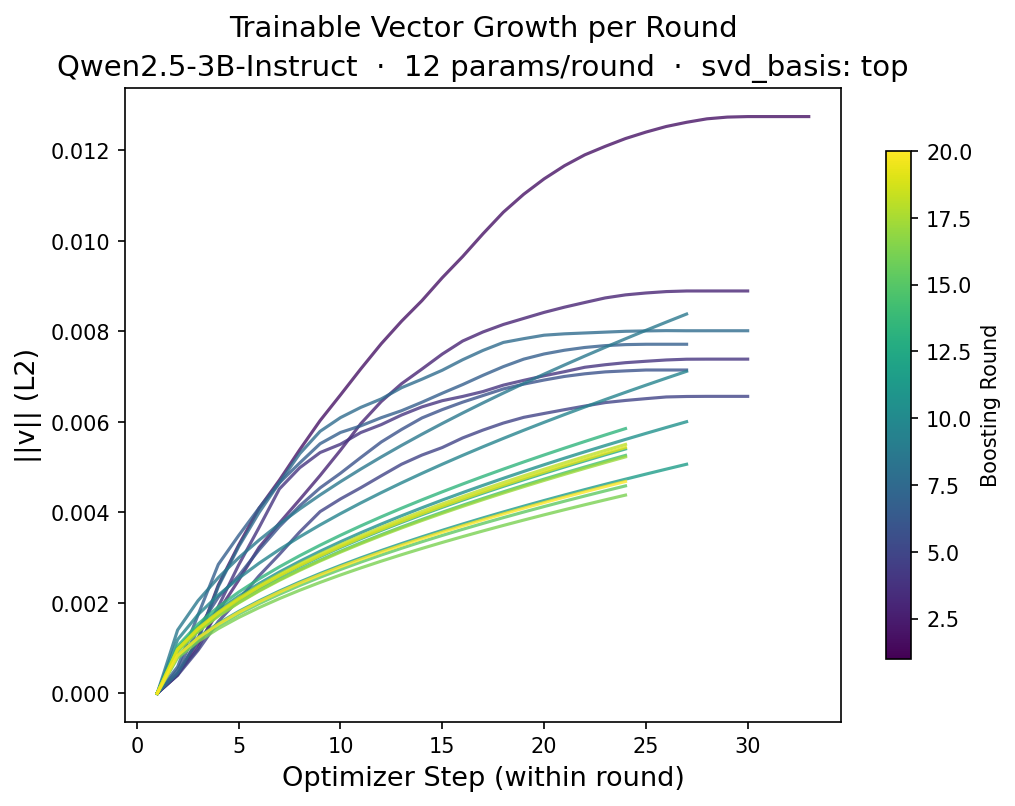}
    \caption{\textbf{Boosting dynamics on GSM8K with \textsc{top}.}
    Left: training accuracy rises from 90.8\% to 93.8\% as failures fall from
    687 to 462. Right: later rounds converge to smaller trainable-vector norms
    as the failure set shrinks.}
    \label{fig:dynamics}
\end{figure}

\paragraph{Boosting dynamics.}
Figure~\ref{fig:dynamics} shows that failures decrease from 687 to 462 over
20 rounds while training accuracy rises from 90.8\% to 93.8\%. Despite
approximately 2\% average regression of previously correct examples per round,
held-out accuracy increases from 76.0\% to 89.1\%, narrowing the train-test
gap from 14.8 to 4.7 points and providing no evidence of failure-set memorization.
Adapter norms decline from $\approx 0.013$ in early rounds to 0.005 in later
rounds, and later rounds require fewer optimizer steps (24 vs 33); the same
patterns are qualitatively observed with \textsc{rotate}.

\section{Conclusion}
\label{sec:conclusion}
BoostLoRA overcomes the expressivity limit of ultra-low-parameter adapters by sequentially training and merging them on failure sets, with a \textsc{rotate} basis that grows effective rank to $rT$.
It surpasses both TinyLoRA and full fine-tuning on math reasoning (89.1\% on GSM8K with 12 parameters per round vs.\ 87.0\% with 3.09B), avoids catastrophic degradation seen in full fine-tuning narrow datasets (HumanEval: 80.4\% vs.\ 57.9\%), and transfers to protein binding classification on an encoder-only pLM.
Boosting is most useful when individual adapters are weak; larger per-round budgets can overcorrect and require learning-rate scaling. The approach trades peak parameter efficiency for sequential, non-parallel rounds and full-dataset failure scans, and remains untested beyond Qwen2.5-3B, ESM2-650M, and the evaluated datasets.

\newpage


\subsection*{AI use statement}
In this work, we used generative AI tools solely for copy-editing of the
manuscript-grammar, syntax, and clarity suggestions on author-written prose.
We have not used generative AI tools for research ideation, the theoretical
statements and proofs, experimental design, implementation, data analysis, or
the conclusions; these were produced and verified by the authors. We have
reviewed all AI-assisted edits. We take full responsibility for the final
content of this work, including all text, claims, and artifacts.

\subsection*{Ethics statement}
This work uses only publicly available models (Qwen2.5-3B-Instruct, ESM2-650M)
and public benchmark datasets (GSM8K, MATH-500, MBPP, HumanEval, and
PPB-Affinity). It involves no human subjects, no personally identifiable
information, and releases no new data. We are not aware of ethical concerns
raised by the method or its evaluation beyond the general considerations that
apply to parameter-efficient fine-tuning of large models.

\subsection*{Reproducibility statement}
The BoostLoRA procedure is fully specified in Algorithm~\ref{alg:boostlora} and
Section~\ref{sec:method}, including the TinyLoRA parameterization
(\S\ref{sec:adapter}), the \textsc{rotate} SVD basis (\S\ref{sec:rank}), and the
failure-set construction, training procedure, and stopping rules
(\S\ref{sec:grpo}, \S\ref{sec:experiments}). All models, datasets, benchmarks,
baselines, and training hyperparameters are listed in
Section~\ref{sec:experiments}. Complete proofs of Theorems~\ref{thm:rank}
and~\ref{thm:gen} appear in the main text and Appendix~\ref{sec:gen_proof};
end-to-end training costs are reported in Appendix~\ref{sec:training_cost}; the
controlled empirical validation of the theory is in
Appendix~\ref{sec:theory_validation}; and the details of our Chain-of-LoRA
re-implementation are given in Appendix~\ref{sec:cola_comparison}.

\bibliographystyle{iclr2027_conference}
\bibliography{iclr2027_conference}

\clearpage

\appendix

\section{Complexity Comparison}
\label{sec:complexity_appendix}

Table~\ref{tab:complexity} compares peak trainable parameters, cumulative training parameters, and effective rank across PEFT methods. BoostLoRA exposes $\mathcal{O}(gu)$ trainable parameters in each round and optimizes $\mathcal{O}(Tgu)$ distinct adapter coordinates over $T$ rounds. After merging, the final model retains no additional adapter parameters. With \textsc{rotate}, the cumulative update has rank at most $\min(rT,d,k)$ (Theorem~\ref{thm:rank}), decoupling peak trainable state from cumulative representational capacity.

\begin{table}[t]
    \centering
    \caption{\textbf{Parameter complexity comparison.}
    $n$: layers, $m$: modules per layer, $d$: width,
    $r$: rank, $u$: projection dimension.
    All prior methods have fixed effective rank~$r$.
    BoostLoRA's cumulative rank grows
    with~$T$ (up to $\min(rT, d, k)$ with \textsc{rotate}),
    even though each round's adapter is rank~$r$.
    Extends Table~1 of~\citet{tinylora}.}
    \label{tab:complexity}
    \small
    \begin{tabular}{lccc}
        \toprule
        \textbf{Method}
            & \textbf{Training Parameters}
            & \textbf{Minimum}
            & \textbf{Effective Rank} \\
        \midrule
        Full FT        & $\mathcal{O}(nmd^2)$     & $nmd^2$ & $d$ \\
        LoRA           & $\mathcal{O}(nmdr)$      & $2nmd$  & $r$ \\
        LoRA-XS        & $\mathcal{O}(nmr^2)$     & $nm$    & $r$ \\
        VeRA           & $\mathcal{O}(nm(d{+}r))$ & $2nmd$  & $r$ \\
        TinyLoRA       & $\mathcal{O}(nmu)$        & $1$     & $r$ \\
        \midrule
        \textbf{BoostLoRA}
            & \shortstack{$\mathcal{O}(gu)$ peak\\
                  $\mathcal{O}(Tgu)$ cumulative}
                    & $T$ & $\mathbf{\leq rT}$ \\
        \bottomrule
    \end{tabular}
\end{table}

\section{Training Cost}
\label{sec:training_cost}
BoostLoRA introduces sequential training cost because each round requires identifying the current failure set before training a new adapter. Table~\ref{tab:wallclock} reports end-to-end wall-clock time and the measured fractions attributable to failure-set evaluation and adapter training. The reported percentages need not sum to 100\%; the remainder contains held-out evaluation and other pipeline overheads.

\begin{table}[t]
\centering
\small
\caption{\textbf{End-to-end BoostLoRA training cost.}
All experiments use 20 boosting rounds. ``Failure scan'' is the full-training-set evaluation used to construct $\mathcal{F}_t$.}
\label{tab:wallclock}
\setlength{\tabcolsep}{6pt}
\begin{tabular}{lrrr}
\toprule
\textbf{Dataset}
    & \textbf{Wall clock (h)}
    & \textbf{Failure scan}
    & \textbf{Adapter training} \\
\midrule
GSM8K   & 184.2 & 14.4\% & 80.4\% \\
MATH-500 & 257.8 &  4.5\% & 79.2\% \\
MBPP     &  14.0 &  4.5\% & 79.2\% \\
\bottomrule
\end{tabular}
\end{table}

The failure scan is measurable but does not dominate runtime; GRPO rollout generation and optimization account for most of the measured cost. Training becomes cheaper as failures shrink. On GSM8K, per-round runtime decreased from 536.8 minutes in round~1 to 469.5 minutes in round~20, a reduction of 67.3 minutes (12.5\%). The rounds remain sequential and therefore cannot be parallelized across boosting depth.

\section{Proof of Theorem~\ref{thm:gen}}
\label{sec:gen_proof}

We bound the generalization error of the BoostLoRA ensemble after $T$ rounds using Rademacher complexity.

\paragraph{Function class per round.}
At round $t$, the adapter perturbs each adapted layer's output by
$\Delta W_\ell\, h_\ell(x) = U_\ell \Sigma_\ell R_t V_\ell^\top h_\ell(x)$,
where $h_\ell(x)$ is the input to layer~$\ell$.
Under the linearization approximation (treating the network above each
adapted layer as frozen at its pre-adaptation state), the margin
perturbation is
$\Delta m(x) \approx \sum_\ell g_\ell(x)^\top \Delta W_\ell\, h_\ell(x)$,
where $g_\ell(x) \in \mathbb{R}^d$ is the
gradient of the margin with respect to layer~$\ell$'s output.
Substituting $R_t = \sum_i (v_t)_i P_i$ gives
\begin{equation}
  f_t(x) = \langle v_t,\, \phi_t(x) \rangle, \qquad
  \phi_t(x)_i = \sum_\ell g_\ell(x)^\top U_\ell\, \Sigma_\ell\, P_i\, V_\ell^\top h_\ell(x),
\end{equation}
where all quantities except $v_t$ are frozen.
Since $f_t$ is linear in $v_t$, the round-$t$ function class is
$\mathcal{F}_t = \{x \mapsto \langle v_t, \phi_t(x)\rangle : \|v_t\|_2 \leq B_t\}$.

\paragraph{Rademacher complexity of one round.}
The empirical Rademacher complexity of $\mathcal{F}_t$ over $n$ samples $x_1, \ldots, x_n$ is
\begin{align}
  \hat{\mathcal{R}}_n(\mathcal{F}_t)
  &= \frac{1}{n}\, \mathbb{E}_\sigma\!\left[
       \sup_{\|v_t\| \leq B_t} \sum_{j=1}^{n} \sigma_j \langle v_t, \phi_t(x_j)\rangle
     \right] \notag \\
  &= \frac{B_t}{n}\, \mathbb{E}_\sigma\!\left[
       \Bigl\|\textstyle\sum_{j=1}^{n} \sigma_j\, \phi_t(x_j)\Bigr\|_2
     \right],
\end{align}
where $\sigma_j$ are iid Rademacher variables and the supremum is attained at $v_t \propto \sum_j \sigma_j \phi_t(x_j)$.
By Jensen's inequality,
\begin{equation}
  \mathbb{E}_\sigma\!\left[\Bigl\|\textstyle\sum_j \sigma_j\, \phi_t(x_j)\Bigr\|\right]
  \;\leq\; \sqrt{\sum_j \|\phi_t(x_j)\|^2}
  \;\leq\; \sqrt{n}\, \max_x \|\phi_t(x)\|
  \;=\; \sqrt{n}\, X_t,
\end{equation}
so $\hat{\mathcal{R}}_n(\mathcal{F}_t) \leq B_t X_t / \sqrt{n}$.

\paragraph{Subadditivity across rounds.}
The cumulative perturbation after $T$ rounds is $F(x) = \sum_{t=1}^T f_t(x)$, which lives in the Minkowski sum $\mathcal{F}_1 + \cdots + \mathcal{F}_T$.
Rademacher complexity is subadditive over Minkowski sums:
\begin{equation}
\label{eqn:rad}
  \mathcal{R}_n\!\left(\textstyle\sum_t \mathcal{F}_t\right)
  \;\leq\; \sum_t \mathcal{R}_n(\mathcal{F}_t)
  \;\leq\; \sum_t \frac{B_t X_t}{\sqrt{n}}
  \;\leq\; \frac{B_{\mathrm{total}}  X}{\sqrt{n}},
\end{equation}
where $X = \max_t X_t$ and $B_{\mathrm{total}} = \sum_t B_t$.

When using the \textsc{Top} basis, the Rademacher complexity matches the general bound of Eq.~\ref{eqn:rad}. Every round uses the same top-$r$ singular vectors, so $\phi_t = \phi_1$ for all $t$ and $X_t = X_1 = \max_x \|\phi_1(x)\|_2$. With uniform $B_t = B$ the subadditive bound above becomes $\mathcal{R}_n \leq B X_1 T / \sqrt{n}$, which is linear in $T$. 

When using the \textsc{rotate} basis, the Rademacher complexity is smaller than Eq.~\ref{eqn:rad} by a factor of up to $\sqrt{T}$. Round $t$ projects $g_\ell(x)$ and $h_\ell(x)$ onto its own block of singular directions, so the $\phi_t(x)$ are orthogonal components of one vector and $H^2 = \max_x \sum_t \|\phi_t(x)\|_2^2$ is bounded independently of $T$. Bounding the concatenated predictor $F(x) = \langle v, \Phi(x) \rangle$ directly, with $v = (v_1, \ldots, v_T)$ and $\Phi(x) = (\phi_1(x), \ldots, \phi_T(x))$, gives $\mathcal{R}_n \leq \sqrt{\sum_t B_t^2}\, H / \sqrt{n}$, which for uniform $B_t = B$ is $B H \sqrt{T} / \sqrt{n}$. Under \textsc{top} the same argument gives $H^2 = T \max_x \|\phi_1(x)\|_2^2$ and recovers $B X_1 T / \sqrt{n}$, so the gain is due to the basis.

\paragraph{Margin bound.}
Applying the standard Rademacher generalization bound~\citep{bartlett_mendelson} (see Theorem~3.3 of~\citet{mohri2018}) to the margin loss $\ell_\theta(x) = \min(1, \max(0, 1 - m(x)/\theta))$, which maps to $[0,1]$: for any margin $\theta > 0$, with probability $\geq 1 - \delta$,
\begin{equation}
  \Pr_{\mathrm{test}}[\mathrm{error}]
  \;\leq\; \Pr_{\mathrm{train}}[m_T(x) \leq \theta]
  \;+\; \frac{2}{\theta}\, \mathcal{R}_n\!\left(\textstyle\sum_t \mathcal{F}_t\right)
  \;+\; \sqrt{\frac{\log(2/\delta)}{2n}},
\end{equation}
which gives the stated bound after substituting the Rademacher estimate from Step~3.

\paragraph{Approximation.}
The proof treats $\phi_t(x)$ as fixed for each round, but in practice $h_\ell(x)$ changes after merging previous adapters. This is a reasonable approximation because the per-round weight perturbation is small ($\|\Delta W_t\|_F \approx 0.01$), so hidden states shift only slightly between rounds. The proof treats $g_\ell(x)$ and $h_\ell(x)$ as fixed at their pre-adaptation values, but both shift after merging previous adapters. The linearization error is $|F(x) - \tilde{F}(x)| = O(T \|\Delta W\|_F B_{\mathrm{total}})$, which is negligible as $\|\Delta W_t\|_F \ll 1$ per round.

\section{Empirical Theory Validation}
\label{sec:theory_validation}

\begin{figure}[t]
    \centering
    \includegraphics[width=\linewidth]{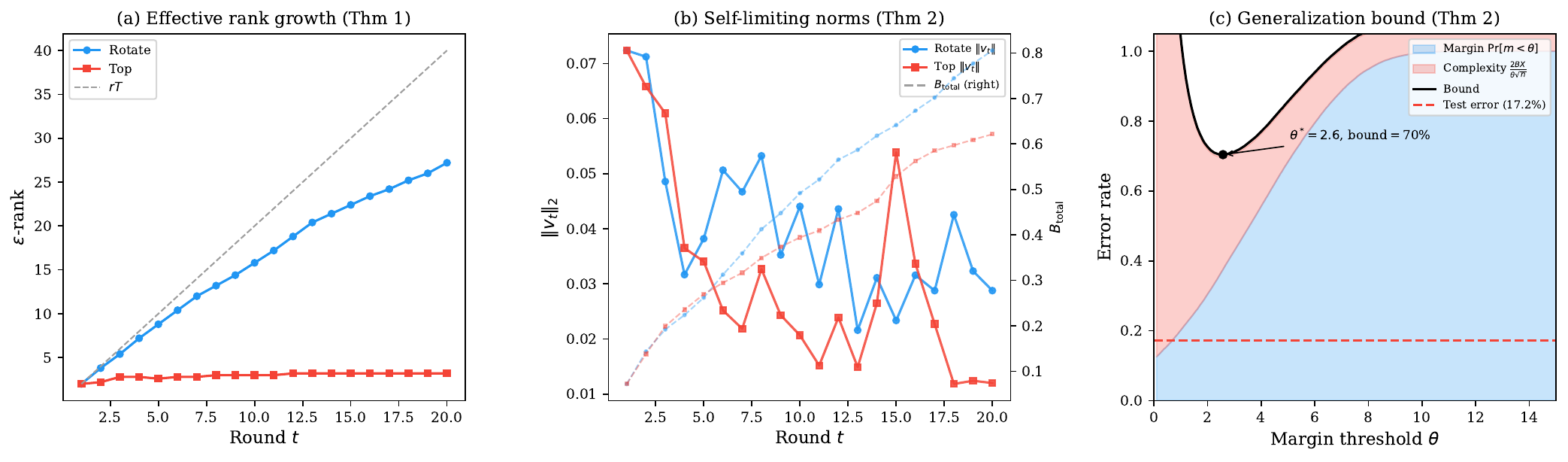}
    \caption{\textbf{Empirical validation of theoretical properties} on CIFAR-10 (CNN, $r{=}2$, $T{=}20$, 9 params/round).
    \textbf{(a)}~$\varepsilon$-rank of the cumulative weight update $\bar{\Delta} = \sum_t \Delta_t$. \textsc{Rotate} (blue) grows linearly, tracking the theoretical bound $rT$. \textsc{Top} (red) saturates at ${\approx}\,3$, confirming Theorem~\ref{thm:rank}.
    \textbf{(b)}~Per-round adapter norms $\|v_t\|_2$ (solid) and cumulative norm $B_{\mathrm{total}} = \sum_{s \leq t} \|v_s\|$ (dashed, right axis). Both bases show declining per-round norms; $B_{\mathrm{total}}$ grows sublinearly, consistent with the self-limiting property in Theorem~\ref{thm:gen}.
    \textbf{(c)}~Generalization bound (Theorem~\ref{thm:gen}) evaluated numerically. The bound (black curve) decomposes into a margin term (blue, fraction of training examples with margin below $\theta$) and a complexity term (red, $\propto B_{\mathrm{total}} \cdot X / \theta\sqrt{n}$). At the optimal $\theta^* = 2.6$, the bound is 70\%, non-vacuous against the observed 17.2\% test error.}
    \label{fig:theory_validation}
\end{figure}

We validate Theorems~\ref{thm:rank} and~\ref{thm:gen} on a controlled CIFAR-10 classification task using a small CNN with 5 adapted layers, $r{=}2$, $T{=}20$, and 9 trainable parameters per round. This isolates the boosting mechanism from LLM-specific confounders (tokenization, generation, reward noise). The CNN includes LayerNorm before each fully connected layer, matching the normalization structure of transformer architectures.
Figure~\ref{fig:theory_validation} summarizes the results.

The $\varepsilon$-rank of the cumulative update $\bar{\Delta}$ grows linearly with $T$ for \textsc{rotate}, reaching ${\approx}\,27$ at round~20. The gap from the theoretical maximum $rT = 40$ reflects the self-limiting dynamics visible in panel~(b): later rounds produce smaller adapters whose singular values fall below the $\varepsilon$-rank threshold, even though the subspaces are orthogonal. On Qwen2.5-3B, where adapter norms are more uniform across rounds, $\varepsilon$-rank tracks $rT$ exactly (Figure~\ref{fig:rank_comparison}). With \textsc{top}, $\varepsilon$-rank stays at ${\approx}\,3$ regardless of $T$, confirming that successive adapters overwrite the same subspace.

Per-round adapter norms decline over boosting rounds for both bases: from ${\approx}\,0.07$ in early rounds to ${\approx}\,0.01{-}0.03$ in later rounds. \textsc{Top} norms decay faster because the fixed subspace offers diminishing returns. The cumulative norm $B_{\mathrm{total}}$ grows sublinearly, consistent with the convergence argument in Theorem~\ref{thm:gen}. As the failure set shrinks, the gradient signal weakens, producing smaller adapters.

We evaluate Theorem~\ref{thm:gen} numerically using the measured quantities: $B_{\mathrm{total}} = 0.81$, feature norm $X = 115.6$ (computed from hidden-state activations at the adapted layers), $n = 50{,}000$, and the empirical margin distribution at round~20. The bound is minimized at $\theta^* = 2.6$, where the margin term (37.5\%) and complexity term (32.3\%) balance. The resulting bound of 70\% is non-vacuous, though loose relative to the observed 17.2\% test error. The looseness is driven by $X$: in architectures with smaller weight singular values or stronger normalization, $X$ decreases and the bound tightens proportionally.

\section{Chain of LoRA Comparison}
\label{sec:cola_comparison}
We implemented Chain of LoRA (CoLA) from the description in the original paper,
as no public repository is available. We mirrored BoostLoRA's use of Qwen2.5-3B
in both its base and instruct variants, LoRA rank $r{=}8$ and $\alpha{=}16$ on
the \texttt{q\_proj} and \texttt{v\_proj} projections, and a batch size of 8
(reduced only for MATH-500 to avoid OOM). Train/test splits were held identical
to the BoostLoRA experiments. We ran five random seeds per configuration and
report the best. We evaluated chain lengths
$t \in \{1, 2, 3, 8, 10, 15, 20, 25, 30\}$, where $t{=}1$ recovers vanilla LoRA
and each subsequent stage trains a new adapter on the residual and merges it into
the running weights. Training used the L4 GPU (the original CoLA experiments used
A100s; we were constrained by availability). The learning rate was selected by
grid search around the range reported in the CoLA paper. Full per-seed-best
results are in Table~\ref{tab:chain-of-lora}.

\begin{table}[h]
\centering
\caption{Chain of LoRA accuracy across chain lengths $t$. 
$t{=}1$ is vanilla LoRA.}
\label{tab:chain-of-lora}
\resizebox{\textwidth}{!}{%
\begin{tabular}{lccccccccc}
\toprule
Dataset & $t{=}1$ & $t{=}2$ & $t{=}3$ & $t{=}8$ & $t{=}10$ & $t{=}15$ & $t{=}20$ & $t{=}25$ & $t{=}30$ \\
\midrule
MBPP (base)          & 0.498  & \textbf{0.526} & 0.514  & 0.328  & 0.342  & 0.316          & 0.298   & 0.292  & 0.248  \\
MBPP (instruct)      & \textbf{0.502} & 0.450  & 0.458  & 0.492  & 0.488  & \textbf{0.502} & 0.458   & 0.438  & 0.402  \\
HumanEval (base)     & 0.4146 & \textbf{0.4817} & 0.4634 & 0.2866 & 0.2866 & 0.3171        & 0.2988  & 0.2134 & 0.2073 \\
HumanEval (instruct) & \textbf{0.7500} & 0.7195 & 0.7134 & 0.7073 & 0.7134 & 0.7073       & 0.6890  & 0.6707 & 0.6646 \\
GSM8K (base)         & 0.2684 & \textbf{0.3427} & 0.2782 & 0.2934 & 0.2881 & 0.2964        & 0.2472  & 0.2252 & 0.2199 \\
GSM8K (instruct)     & 0.0591 & 0.0614 & 0.0569 & 0.0697 & 0.0637 & \textbf{0.0735} & 0.0773 & 0.0538 & 0.0803 \\
\bottomrule
\end{tabular}%
}

\end{table}

\end{document}

%% file: math_commands.tex
\usepackage{amsmath,amsfonts,bm}

\def\eqref#1{equation~\ref{#1}}

\def\1{\bm{1}}

\DeclareMathAlphabet{\mathsfit}{\encodingdefault}{\sfdefault}{m}{sl}
\SetMathAlphabet{\mathsfit}{bold}{\encodingdefault}{\sfdefault}{bx}{n}



%% file: iclr2027_conference.bib
@misc{tinylora,
      title={Learning to Reason in 13 Parameters},
      author={John X. Morris and Niloofar Mireshghallah and Mark Ibrahim and Saeed Mahloujifar},
      year={2026},
      eprint={2602.04118},
      archivePrefix={arXiv},
      primaryClass={cs.LG},
      url={https://arxiv.org/abs/2602.04118},
}

@misc{grpo,
      title={DeepSeekMath: Pushing the Limits of Mathematical Reasoning in Open Language Models},
      author={Zhihong Shao and Peiyi Wang and Qihao Zhu and Runxin Xu and Junxiao Song and Xiao Bi and Haowei Zhang and Mingchuan Zhang and Y. K. Li and Y. Wu and Daya Guo},
      year={2024},
      eprint={2402.03300},
      archivePrefix={arXiv},
      primaryClass={cs.CL},
      url={https://arxiv.org/abs/2402.03300},
}

@misc{qwen2.5,
    title = {Qwen2.5: A Party of Foundation Models},
    url = {https://qwenlm.github.io/blog/qwen2.5/},
    author = {QwenTeam},
    month = {September},
    year = {2024}
}

@misc{relora,
      title={ReLoRA: High-Rank Training Through Low-Rank Updates},
      author={Vladislav Lialin and Namrata Shivagunde and Sherin Muckatira and Anna Rumshisky},
      year={2023},
      eprint={2307.05695},
      archivePrefix={arXiv},
      primaryClass={cs.CL},
      url={https://arxiv.org/abs/2307.05695},
}

@article{gsm8k,
  title={Training Verifiers to Solve Math Word Problems},
  author={Cobbe, Karl and Kosaraju, Vineet and Bavarian, Mohammad and Chen, Mark and Jun, Heewoo and Kaiser, Lukasz and Plappert, Matthias and Tworek, Jerry and Hilton, Jacob and Nakano, Reiichiro and Hesse, Christopher and Schulman, John},
  journal={arXiv preprint arXiv:2110.14168},
  year={2021}
}

@article{math500,
      title={Let's Verify Step by Step},
      author={Lightman, Hunter and Kosaraju, Vineet and Burda, Yura and Edwards, Harri and Baker, Bowen and Lee, Teddy and Leike, Jan and Schulman, John and Sutskever, Ilya and Cobbe, Karl},
      journal={arXiv preprint arXiv:2305.20050},
      year={2023}
}

@article{mbpp,
  title={Program Synthesis with Large Language Models},
  author={Austin, Jacob and Odena, Augustus and Nye, Maxwell and Bosma, Maarten and Michalewski, Henryk and Dohan, David and Jiang, Ellen and Cai, Carrie and Terry, Michael and Le, Quoc and others},
  journal={arXiv preprint arXiv:2108.07732},
  year={2021}
}

@misc{humaneval,
      title={Evaluating Large Language Models Trained on Code},
      author={Mark Chen and Jerry Tworek and Heewoo Jun and Qiming Yuan and Henrique Ponde de Oliveira Pinto and Jared Kaplan and Harri Edwards and Yuri Burda and Nicholas Joseph and Greg Brockman and Alex Ray and Raul Puri and Gretchen Krueger and Michael Petrov and Heidy Khlaaf and Girish Sastry and Pamela Mishkin and Brooke Chan and Scott Gray and Nick Ryder and Mikhail Pavlov and Alethea Power and Lukasz Kaiser and Mohammad Bavarian and Clemens Winter and Philippe Tillet and Felipe Petroski Such and Dave Cummings and Matthias Plappert and Fotios Chantzis and Elizabeth Barnes and Ariel Herbert-Voss and William Hebgen Guss and Alex Nichol and Alex Paino and Nikolas Tezak and Jie Tang and Igor Babuschkin and Suchir Balaji and Shantanu Jain and William Saunders and Christopher Hesse and Andrew N. Carr and Jan Leike and Josh Achiam and Vedant Misra and Evan Morikawa and Alec Radford and Matthew Knight and Miles Brundage and Mira Murati and Katie Mayer and Peter Welinder and Bob McGrew and Dario Amodei and Sam McCandlish and Ilya Sutskever and Wojciech Zaremba},
      year={2021},
      eprint={2107.03374},
      archivePrefix={arXiv},
      primaryClass={cs.LG}
}

@misc{stepcoder,
      title={StepCoder: Improve Code Generation with Reinforcement Learning from Compiler Feedback},
      author={Shihan Dou and Yan Liu and Haoxiang Jia and Limao Xiong and Enyu Zhou and Wei Shen and Junjie Shan and Caishuang Huang and Xiao Wang and Xiaoran Fan and Zhiheng Xi and Yuhao Zhou and Tao Ji and Rui Zheng and Qi Zhang and Xuanjing Huang and Tao Gui},
      year={2024},
      eprint={2402.01391},
      archivePrefix={arXiv},
      primaryClass={cs.SE},
      url={https://arxiv.org/abs/2402.01391},
}

@misc{mucode,
      title={Multi-Turn Code Generation Through Single-Step Rewards},
      author={Arnav Kumar Jain and Gonzalo Gonzalez-Pumariega and Wayne Chen and Alexander M Rush and Wenting Zhao and Sanjiban Choudhury},
      year={2025},
      eprint={2502.20380},
      archivePrefix={arXiv},
      primaryClass={cs.LG},
      url={https://arxiv.org/abs/2502.20380},
}

@misc{simplerl,
      title={SimpleRL-Zoo: Investigating and Taming Zero Reinforcement Learning for Open Base Models in the Wild},
      author={Weihao Zeng and Yuzhen Huang and Qian Liu and Wei Liu and Keqing He and Zejun Ma and Junxian He},
      year={2025},
      eprint={2503.18892},
      archivePrefix={arXiv},
      primaryClass={cs.LG},
      url={https://arxiv.org/abs/2503.18892},
}

@misc{lora,
      title={LoRA: Low-Rank Adaptation of Large Language Models},
      author={Edward J. Hu and Yelong Shen and Phillip Wallis and Zeyuan Allen-Zhu and Yuanzhi Li and Shean Wang and Lu Wang and Weizhu Chen},
      year={2021},
      eprint={2106.09685},
      archivePrefix={arXiv},
      primaryClass={cs.CL},
      url={https://arxiv.org/abs/2106.09685},
}

@misc{loraxs,
      title={LoRA-XS: Low-Rank Adaptation with Extremely Small Number of Parameters},
      author={Klaudia Bałazy and Mohammadreza Banaei and Karl Aberer and Jacek Tabor},
      year={2025},
      eprint={2405.17604},
      archivePrefix={arXiv},
      primaryClass={cs.LG},
      url={https://arxiv.org/abs/2405.17604},
}

@misc{adalora,
      title={AdaLoRA: Adaptive Budget Allocation for Parameter-Efficient Fine-Tuning},
      author={Qingru Zhang and Minshuo Chen and Alexander Bukharin and Nikos Karampatziakis and Pengcheng He and Yu Cheng and Weizhu Chen and Tuo Zhao},
      year={2023},
      eprint={2303.10512},
      archivePrefix={arXiv},
      primaryClass={cs.CL},
      url={https://arxiv.org/abs/2303.10512},
}

@misc{pissa,
      title={PiSSA: Principal Singular Values and Singular Vectors Adaptation of Large Language Models},
      author={Fanxu Meng and Zhaohui Wang and Muhan Zhang},
      year={2025},
      eprint={2404.02948},
      archivePrefix={arXiv},
      primaryClass={cs.LG},
      url={https://arxiv.org/abs/2404.02948},
}

@misc{vera,
      title={VeRA: Vector-based Random Matrix Adaptation},
      author={Dawid J. Kopiczko and Tijmen Blankevoort and Yuki M. Asano},
      year={2024},
      eprint={2310.11454},
      archivePrefix={arXiv},
      primaryClass={cs.CL},
      url={https://arxiv.org/abs/2310.11454},
}

@misc{rlvr_peft,
      title={Evaluating Parameter Efficient Methods for RLVR},
      author={Qingyu Yin and Yulun Wu and Zhennan Shen and Sunbowen Li and Zhilin Wang and Yanshu Li and Chak Tou Leong and Jiale Kang and Jinjin Gu},
      year={2025},
      eprint={2512.23165},
      archivePrefix={arXiv},
      primaryClass={cs.LG},
      url={https://arxiv.org/abs/2512.23165},
}

@ARTICLE{friedman2001,
       author = {{Friedman}, Jerome H.},
        title = "{Greedy function approximation: A gradient boosting machine.}",
      journal = {Annals of Statistics},
         year = 2001,
        month = oct,
       volume = {29},
       number = {5},
        pages = {1189--1232},
          doi = {10.1214/aos/1013203451},
       adsurl = {https://ui.adsabs.harvard.edu/abs/2001AnSta..2903451F}
}

@book{mohri2018,
  title={Foundations of machine learning},
  author={Mohri, Mehryar and Rostamizadeh, Afshin and Talwalkar, Ameet},
  year={2018},
  publisher={MIT press}
}

@article{bartlett_mendelson,
  title={Rademacher and Gaussian complexities: Risk bounds and structural results},
  author={Bartlett, Peter L. and Mendelson, Shahar},
  journal={Journal of Machine Learning Research},
  volume={3},
  pages={463--482},
  year={2002}
}

@misc{dora,
      title={DoRA: Weight-Decomposed Low-Rank Adaptation},
      author={Shih-Yang Liu and Chien-Yi Wang and Hongxu Yin and Pavlo Molchanov and Yu-Chiang Frank Wang and Kwang-Ting Cheng and Min-Hung Chen},
      year={2024},
      eprint={2402.09353},
      archivePrefix={arXiv},
      primaryClass={cs.CL},
      url={https://arxiv.org/abs/2402.09353},
}

@misc{cola,
      title={Chain of LoRA: Efficient Fine-tuning of Language Models via Residual Learning},
      author={Wenhan Xia and Chengwei Qin and Elad Hazan},
      year={2024},
      eprint={2401.04151},
      archivePrefix={arXiv},
      primaryClass={cs.LG},
      url={https://arxiv.org/abs/2401.04151},
}

@inproceedings{melora,
    title = "{MEL}o{RA}: Mini-Ensemble Low-Rank Adapters for Parameter-Efficient Fine-Tuning",
    author = "Ren, Pengjie  and
      Shi, Chengshun  and
      Wu, Shiguang  and
      Zhang, Mengqi  and
      Ren, Zhaochun  and
      de Rijke, Maarten  and
      Chen, Zhumin  and
      Pei, Jiahuan",
    editor = "Ku, Lun-Wei  and
      Martins, Andre  and
      Srikumar, Vivek",
    booktitle = "Proceedings of the 62nd Annual Meeting of the Association for Computational Linguistics (Volume 1: Long Papers)",
    month = aug,
    year = "2024",
    address = "Bangkok, Thailand",
    publisher = "Association for Computational Linguistics",
    url = "https://aclanthology.org/2024.acl-long.168/",
    doi = "10.18653/v1/2024.acl-long.168",
    pages = "3052--3064"
}

@misc{xgblora,
      title={Less is More: Extreme Gradient Boost Rank-1 Adaption for Efficient Finetuning of LLMs},
      author={Yifei Zhang and Hao Zhu and Aiwei Liu and Han Yu and Piotr Koniusz and Irwin King},
      year={2024},
      eprint={2410.19694},
      archivePrefix={arXiv},
      primaryClass={cs.CL},
      url={https://arxiv.org/abs/2410.19694},
}

@misc{modelsoups,
      title={Model soups: averaging weights of multiple fine-tuned models improves accuracy without increasing inference time},
      author={Mitchell Wortsman and Gabriel Ilharco and Samir Yitzhak Gadre and Rebecca Roelofs and Raphael Gontijo-Lopes and Ari S. Morcos and Hongseok Namkoong and Ali Farhadi and Yair Carmon and Simon Kornblith and Ludwig Schmidt},
      year={2022},
      eprint={2203.05482},
      archivePrefix={arXiv},
      primaryClass={cs.LG},
      url={https://arxiv.org/abs/2203.05482},
}

@misc{lorasoups,
      title={LoRA Soups: Merging LoRAs for Practical Skill Composition Tasks},
      author={Akshara Prabhakar and Yuanzhi Li and Karthik Narasimhan and Sham Kakade and Eran Malach and Samy Jelassi},
      year={2024},
      eprint={2410.13025},
      archivePrefix={arXiv},
      primaryClass={cs.CL},
      url={https://arxiv.org/abs/2410.13025},
}

@article{deepseekr1,
   title={DeepSeek-R1 incentivizes reasoning in LLMs through reinforcement learning},
   volume={645},
   ISSN={1476-4687},
   url={http://dx.doi.org/10.1038/s41586-025-09422-z},
   DOI={10.1038/s41586-025-09422-z},
   number={8081},
   journal={Nature},
   publisher={Springer Science and Business Media LLC},
   author={Guo, Daya and Yang, Dejian and Zhang, Haowei and Song, Junxiao and Wang, Peiyi and Zhu, Qihao and Xu, Runxin and Zhang, Ruoyu and Ma, Shirong and Bi, Xiao and others},
   year={2025},
   month=sep, pages={633-638} }

@inproceedings{rloo,
    title = "Back to Basics: Revisiting {REINFORCE}-Style Optimization for Learning from Human Feedback in {LLM}s",
    author = {Ahmadian, Arash  and
      Cremer, Chris  and
      Gall{\'e}, Matthias  and
      Fadaee, Marzieh  and
      Kreutzer, Julia  and
      Pietquin, Olivier  and
      {\"U}st{\"u}n, Ahmet  and
      Hooker, Sara},
    editor = "Ku, Lun-Wei  and
      Martins, Andre  and
      Srikumar, Vivek",
    booktitle = "Proceedings of the 62nd Annual Meeting of the Association for Computational Linguistics (Volume 1: Long Papers)",
    month = aug,
    year = "2024",
    address = "Bangkok, Thailand",
    publisher = "Association for Computational Linguistics",
    url = "https://aclanthology.org/2024.acl-long.662/",
    doi = "10.18653/v1/2024.acl-long.662",
    pages = "12248--12267"
}

@article{adaboost,
title = {A Decision-Theoretic Generalization of On-Line Learning and an Application to Boosting},
journal = {Journal of Computer and System Sciences},
volume = {55},
number = {1},
pages = {119-139},
year = {1997},
issn = {0022-0000},
doi = {https://doi.org/10.1006/jcss.1997.1504},
url = {https://www.sciencedirect.com/science/article/pii/S002200009791504X},
author = {Yoav Freund and Robert E Schapire}
}

@inproceedings{schapire1997margin,
  title={Boosting the margin: A new explanation for the effectiveness of voting methods},
  author={Robert E. Schapire and Yoav Freund and Peter Bartlett and Wee Sun Lee},
  booktitle={International Conference on Machine Learning},
  year={1997},
  url={https://api.semanticscholar.org/CorpusID:573509}
}

@inproceedings{xgboost,
author = {Chen, Tianqi and Guestrin, Carlos},
title = {XGBoost: A Scalable Tree Boosting System},
year = {2016},
isbn = {9781450342322},
publisher = {Association for Computing Machinery},
address = {New York, NY, USA},
url = {https://doi.org/10.1145/2939672.2939785},
doi = {10.1145/2939672.2939785},
booktitle = {Proceedings of the 22nd ACM SIGKDD International Conference on Knowledge Discovery and Data Mining},
pages = {785-794},
numpages = {10},
location = {San Francisco, California, USA},
series = {KDD '16}
}

@InProceedings{boostresnet,
  title = 	 {Learning Deep {R}es{N}et Blocks Sequentially using Boosting Theory},
  author =       {Huang, Furong and Ash, Jordan and Langford, John and Schapire, Robert},
  booktitle = 	 {Proceedings of the 35th International Conference on Machine Learning},
  pages = 	 {2058--2067},
  year = 	 {2018},
  editor = 	 {Dy, Jennifer and Krause, Andreas},
  volume = 	 {80},
  series = 	 {Proceedings of Machine Learning Research},
  month = 	 {10--15 Jul},
  publisher =    {PMLR},
  url = 	 {https://proceedings.mlr.press/v80/huang18b.html}
}

@article{esm2,
	author = {Lin, Zeming and Akin, Halil and Rao, Roshan and Hie, Brian and Zhu, Zhongkai and Lu, Wenting and Smetanin, Nikita and Verkuil, Robert and Kabeli, Ori and Shmueli, Yaniv and dos Santos Costa, Allan and Fazel-Zarandi, Maryam and Sercu, Tom and Candido, Salvatore and Rives, Alexander},
	title = {Evolutionary-scale prediction of atomic level protein structure with a language model},
	elocation-id = {2022.07.20.500902},
	year = {2022},
	doi = {10.1101/2022.07.20.500902},
	publisher = {Cold Spring Harbor Laboratory},
	URL = {https://www.biorxiv.org/content/early/2022/10/31/2022.07.20.500902},
	eprint = {https://www.biorxiv.org/content/early/2022/10/31/2022.07.20.500902.full.pdf},
	journal = {bioRxiv}
}

@article{ppb_affinity,
author = {Liu, Huaqing and Chen, Peiyi and Zhai, Xiaochen and Huo, Ku-Geng and Zhou, Shuxian and Han, Lanqing and Fan, Guoxin},
year = {2024},
month = {12},
pages = {},
title = {PPB-Affinity: Protein-Protein Binding Affinity dataset for AI-based protein drug discovery},
volume = {11},
journal = {Scientific Data},
doi = {10.1038/s41597-024-03997-4}
}

@misc{plm_ppi,
      title={Beyond Simple Concatenation: Fairly Assessing PLM Architectures for Multi-Chain Protein-Protein Interactions Prediction},
      author={Hazem Alsamkary and Mohamed Elshaffei and Mohamed Soudy and Sara Ossman and Abdallah Amr and Nehal Adel Abdelsalam and Mohamed Elkerdawy and Ahmed Elnaggar},
      year={2025},
      eprint={2505.20036},
      archivePrefix={arXiv},
      primaryClass={cs.LG},
      url={https://arxiv.org/abs/2505.20036},
}

@inproceedings{NEURIPS2024_cdd06402,
 author = {Yuan, Shen and Liu, Haotian and Xu, Hongteng},
 booktitle = {Advances in Neural Information Processing Systems},
 doi = {10.52202/079017-3606},
 editor = {A. Globerson and L. Mackey and D. Belgrave and A. Fan and U. Paquet and J. Tomczak and C. Zhang},
 pages = {113484--113518},
 publisher = {Curran Associates, Inc.},
 title = {Bridging The Gap between Low-rank and Orthogonal Adaptation via Householder Reflection Adaptation},
 url = {https://proceedings.neurips.cc/paper_files/paper/2024/file/cdd0640218a27e9e2c0e52e324e25db0-Paper-Conference.pdf},
 volume = {37},
 year = {2024}
}

@inproceedings{wang-etal-2023-orthogonal,
    title = "Orthogonal Subspace Learning for Language Model Continual Learning",
    author = "Wang, Xiao  and
      Chen, Tianze  and
      Ge, Qiming  and
      Xia, Han  and
      Bao, Rong  and
      Zheng, Rui  and
      Zhang, Qi  and
      Gui, Tao  and
      Huang, Xuanjing",
    editor = "Bouamor, Houda  and
      Pino, Juan  and
      Bali, Kalika",
    booktitle = "Findings of the Association for Computational Linguistics: EMNLP 2023",
    month = dec,
    year = "2023",
    address = "Singapore",
    publisher = "Association for Computational Linguistics",
    url = "https://aclanthology.org/2023.findings-emnlp.715/",
    doi = "10.18653/v1/2023.findings-emnlp.715",
    pages = "10658--10671"
}
